\documentclass[12pt]{article}
\usepackage{arxiv}
\usepackage[utf8]{inputenc}
\usepackage[T1]{fontenc}
\usepackage[greek,english]{babel}
\usepackage{amsmath}
\usepackage{hyperref}
\usepackage{url}
\usepackage{booktabs}
\usepackage{amsfonts}
\usepackage{graphicx}
\usepackage{subfigure}
\usepackage{natbib}
\usepackage{multirow,enumitem}
\usepackage{array}
\usepackage{algorithm}
\usepackage{authblk}
\usepackage[noend]{algpseudocode}

\newcolumntype{C}[1]{>{\centering\let\newline\\\arraybackslash\hspace{0pt}}m{#1}}

\title{Forging GEMs: Advancing Greek NLP through Quality-Based Corpus Curation}

\author[1]{Alexandra Apostolopoulou}
\author[1]{Konstantinos Kanaris}
\author[1]{Athanasios Koursaris}
\author[1]{George Domalis}
\author[1]{Dimitris Tsakalidis}
\author[1,2]{\href{https://orcid.org/0000-0002-3996-3301}{Ioannis E. Livieris}}

\affil[1]{Novelcore\\
Athens, GR 10436\\
{\small \texttt{\{apostolopoulou,kanaris,koursaris,domalis,tsakalidis,livieris\}@novelcore.eu}}}

\affil[2]{Department of Statistics \& Insurance\\
University of Piraeus\\
Patras, GR 10587\\
{\small \texttt{livieris@unipi.gr}}}

\renewcommand{\shorttitle}{\normalsize{Forging GEMs: Advancing Greek NLP through Quality-Based Corpus Curation}}
\hypersetup{
	pdftitle={Forging GEMs: Advancing Greek NLP through Quality-Based Corpus Curation},
	pdfsubject={cs.CL, cs.AI},
	pdfauthor={Alexandra Apostolopoulou, Konstantinos Kanaris, Athanasios Koursaris, Dimitris Tsakalidis, George Domalis, Ioannis E. Livieris},
	pdfkeywords={Transformer architectures, Legal language modeling, RoBERTa, ELECTRA, ConvBERT, Longformer},
}

\begin{document}
\maketitle

\begin{abstract}
The advancement of natural language processing for morphologically rich and moderately-resourced languages like Modern Greek has been hindered by architectural stagnation, data scarcity, and limited context processing capabilities, particularly in specialized domains such as law. In this work, we propose the Greek Embedding Models (GEMs), a new family of transformer-based language models, specifically developed to address these limitations through architectural diversity and enhanced data curation. The proposed family of models are trained on several large-scale, meticulously curated corpora, encompassing both comprehensive general-domain datasets and specialized legal collections, addressing the persistent data scarcity that has impeded Greek language modeling advancement. The proposed quality-based corpus curation methodology incorporates extensive preprocessing pipelines, sophisticated deduplication strategies and targeted repetition of high-quality legal sub-corpora to enhance domain adaptation. The GEMs family comprises both established architectures (RoBERTa and Longformer) and advanced models not previously applied to Greek (ELECTRA, ConvBERT, and ModernBERT), providing comprehensive coverage of modern transformer designs. Additionally, we introduce the first bilingual Greek-English embedding models tailored for cross-lingual legal applications. Comprehensive evaluation across three core natural language understanding benchmarks demonstrates that the proposed GEM-RoBERTa and GEM-ConvBERT achieve statistically significant performance improvements over established state-of-the-art models, with accuracy gains of up to 3.6\% while conducted statistical analysis using Friedman Aligned-Ranks and Finner post-hoc tests confirms the superiority of our approach across multiple evaluation metrics.
\end{abstract}

% keywords can be removed

\section{Introduction}\label{Sec:introduction}

The advent of transformer-based language models \cite{vaswani2017attention, devlin2019bert} has revolutionized Natural Language Processing (NLP), establishing a dominant paradigm of large-scale pre-training followed by task-specific fine-tuning. While foundational models were initially developed for English, subsequent research has consistently shown that performance is significantly enhanced by creating monolingual models for other languages and adapting them to high-value domains, such as law (e.g., for Greek \cite{papaloukas2021multi, apostolopoulou2022nlp} and German \cite{Darji_2023}) or biomedicine (e.g., for French \cite{labrak2023drbertrobustpretrainedmodel}).
For moderately-resourced languages such as Greek, this specialization is crucial, as dedicated monolingual models like Greek-BERT \cite{koutsikakis2020greekbert} have proven superior to their multilingual counterparts on a range of NLP tasks.

The Greek NLP ecosystem has since evolved rapidly along several parallel tracks. Following the foundational Greek-BERT \cite{koutsikakis2020greekbert}, a line of domain-specific models for the legal field emerged, including the two versions of Greek Legal BERT \cite{papaloukas2021multi, apostolopoulou2022nlp} and the architecturally superior Greek-Legal RoBERTa \cite{saketos2024greeklegalroberta}, which progressively advanced the state-of-the-art on Greek legal tasks. Concurrently, separate efforts addressed this limitation for general-domain Greek with models like the Greek Longformer, which supports a 4096-token context window \cite{zaikisgreeklongformer} and Meltemi-7B \cite{voukoutis2024meltemiopenlargelanguage}, a large-scale generative model with an 8192-token context. However, the divergent progress, has created a fragmented landscape where the benefits of modern architectures and extended context lengths have not been systematically unified for specialized domains like Greek law.

Despite this progress, further advancement in Greek NLP is hindered by several fundamental limitations. Firstly, the development of high-quality, large-scale datasets remains a significant challenge, often leading to sub-optimal model performance. Secondly, researchers also frequently neglect the inherent linguistic complexity of morphologically rich languages like Greek, where intricate syntactic structures and domain-specific terminology demand special attention. This leads to a third limitation: a lack of architectural innovation. The Greek NLP space and the legal domain in particular, has been confined to early encoder architectures like BERT and RoBERTa, often with a restrictive 512-token context length, while more efficient and powerful models such as ELECTRA \cite{clark2020electra} and DeBERTa \cite{he2020deberta} have not been explored. This architectural stagnation prevents the community from leveraging state-of-the-art advancements effectively.

This work is motivated by the need to systematically address these challenges. We bridge the existing gaps by undertaking the first comprehensive pre-training and evaluation of a diverse suite of modern transformer architectures for the Greek language, with a specific focus on the complex legal domain as a proving ground. We specifically target this domain as its highly structured, terminologically rich and stylistically consistent language serves as a low-noise, high-signal corpus, ideal for effectively capturing the complex grammar and syntax of Greek. We propose a new family of Greek Embedding Models (GEMs), pre-trained from scratch on several meticulously curated corpora, including both a large-scale general-purpose dataset and specialized domain-specific ones. Our approach explores architectural innovations not previously applied to Greek and is the first to systematically train and evaluate a suite of Greek models with an extended 1024-token context window. Furthermore, we investigate data-centric training strategies to enhance model performance. In summary, the main contributions of this work are:
\begin{itemize}
	\item We develop the new family of GEMs, comprising both established (RoBERTa and LongFormer) and advanced architectures for Greek (ELECTRA, ConvBERT and ModernBERT), all pre-trained from scratch on our curated corpora to counter the lack of architectural diversity and provide comprehensive coverage of modern transformer designs.
	\item We address data scarcity for the Greek language by constructing and releasing several large-scale, meticulously curated corpora. These include a comprehensive general-domain dataset alongside multiple specialized datasets from the legal domain.
	\item We investigate data-centric strategies, showing that targeted repetition of high-quality legal sub-corpora is an effective strategy for domain adaptation. We also introduce the first bilingual Greek-English Embedding Models specifically tailored for the legal domain.
	\item  We conduct a comprehensive comparative analysis of our models on a diverse set of downstream tasks, covering both the legal domain (named entity recognition, topic classification) and general language understanding (natural language inference). Our findings demonstrate significant performance improvements over existing approaches, establishing strong new baselines and providing a valuable benchmark for the Greek NLP community.
\end{itemize}

The remainder of this work is organized as follows: Section~\ref{Section:2} reviews related work in Greek language modeling. Section~\ref{Section:3} presents the proposed family of model, focusing on dataset composition, corpora configurations and model architectures. Section~\ref{Section:4} presents the results of the numerical experiments and analysis. Finally, Section~\ref{Section:5} discusses our findings and concludes the paper.

\section{Related Work}\label{Section:2}

The development of effective pre-trained language models for morphologically rich languages has emerged as a fundamental challenge in advancing NLP capabilities beyond English-centric paradigms \cite{rogers2020primer,qiu2020pre}. This challenging task has been primarily driven by two complementary research directions: (i) foundational architectural advancements in transformer-based models and their optimization strategies \cite{vaswani2017attention,devlin2019bert,liu2019roberta} and (ii) the systematic adaptation of these innovations to specific languages and specialized domains \cite{martin2020camembert,chalkidis2020legalbert,beltagy2019scibert}. The first direction encompasses sophisticated developments in model architectures, training objectives and efficiency improvements while the second ones focuses on applying these architectural advances to non-English languages and domain-specific applications, where monolingual and specialized models consistently demonstrate superior performance over their multilingual counterparts \cite{virtanen2019multilingual,kenton2019bert}. In the remainder of this section, we provide a comprehensive review of both areas, highlighting their contributions and limitations which motivate our proposed systematic approach to Greek language model development.

\subsection{Architectural Innovations in Pre-trained Language Models}

The evolution of transformer-based language models has been characterized by continuous architectural refinements aimed at improving training efficiency, model capability and contextual understanding. The foundational work began with the transformer architecture \cite{vaswani2017attention} and BERT's \cite{devlin2019bert} Masked Language Model (MLM) objective, which established the dominant paradigm of large-scale pre-training followed by task-specific fine-tuning. This seminal work demonstrated the effectiveness of bidirectional encoding and self-supervised learning for capturing contextual representations across diverse NLP tasks.

Building upon BERT's foundation, RoBERTa \cite{liu2019roberta} introduced critical optimization improvements to the pre-training methodology, including the removal of the Next Sentence Prediction objective, dynamic masking strategies and extended training regimens. These modifications resulted in substantial performance gains across multiple benchmarks, establishing the importance of training procedure optimization alongside architectural innovation. The success of RoBERTa highlighted that careful attention to training dynamics could yield significant improvements without fundamental architectural changes.

ELECTRA \cite{clark2020electra} introduced a paradigm shift through its sample-efficient Replaced Token Detection (RTD) objective, which trains the model to distinguish between original and replaced tokens rather than predicting masked tokens. This approach enables the model to learn from all input positions rather than only the masked subset, resulting in improved sample efficiency and computational effectiveness. The RTD objective demonstrated particular advantages in scenarios with limited computational resources while maintaining competitive performance across downstream tasks.

DeBERTa \cite{he2020deberta} advanced the field through its disentangled attention mechanism, which separately encodes content and position information before combining them in attention computation. This architectural innovation, combined with enhanced mask decoder and absolute position embedding strategies, yielded substantial improvements in model performance while providing better interpretability of attention patterns. The disentangled approach proved particularly effective for tasks requiring fine-grained understanding of positional relationships and syntactic structures.

The challenge of processing longer sequences was addressed by Longformer \cite{beltagy2020longformer}, which introduced sparse attention patterns to handle extended contexts efficiently. By combining local windowed attention, global attention for task-specific tokens and dilated attention patterns, Longformer successfully scaled transformer models to process documents with thousands of tokens while maintaining computational tractability. This innovation proved crucial for applications requiring document-level understanding and long-range dependency modeling.

ConvBERT \cite{jiang2020convbert} introduced a novel hybrid approach which combines convolutional and self-attention mechanisms to improve efficiency and effectiveness. The model replaces some self-attention heads with cheaper convolutional operations, which capture local dependencies more efficiently while maintaining the global modeling capabilities of attention mechanisms. This architectural innovation reduces computational costs while often achieving superior performance, particularly on tasks requiring fine-grained local pattern recognition alongside global context understanding.

Most recently, ModernBERT \cite{warner2024modernbert} has emerged as a comprehensive modernization of the original BERT architecture, incorporating lessons learned from years of research into transformer optimization. ModernBERT integrates several architectural improvements including GeGLU activation functions, RMSNorm normalization, unpadding techniques for efficiency and optimized attention implementations. The model demonstrates that systematic incorporation of incremental improvements can yield substantial performance gains while maintaining compatibility with existing BERT-based infrastructure, representing a mature synthesis of architectural innovations developed over the transformer era.

\subsection{Greek Language Models and Domain Specialization}

The development of Greek natural language processing capabilities has followed a progressive trajectory from foundational monolingual models to specialized domain-specific applications. Koutsikakis et al. \cite{koutsikakis2020greekbert} conducted the pioneering effort in modern Greek NLP by developing Greek BERT, the first monolingual BERT model specifically for the Greek language. Pre-trained on a comprehensive 29GB corpus comprising general-domain text from diverse sources, Greek BERT established crucial performance baselines and demonstrated superior results on tasks such as named entity recognition and natural language inference compared to multilingual models. This foundational work highlighted the significant benefits of language-specific pre-training for morphologically rich languages, where specialized tokenization and vocabulary development prove essential for capturing linguistic nuances.

The success of Greek BERT catalyzed rapid development in domain-specific applications, particularly within the legal field. Papaloukas et al. \cite{papaloukas2021multi} introduced the first Greek Legal BERT by adapting the original Greek BERT architecture through continued pre-training on a 5GB corpus of legal documents. This domain adaptation strategy demonstrated substantial improvements on legal-specific tasks, establishing the effectiveness of specialized pre-training for capturing domain-specific terminology and discourse patterns. The model's superior performance on legal topic classification and named entity recognition tasks validated the importance of domain-focused training data curation.

Apostolopoulou et al. \cite{apostolopoulou2022nlp} extended this work by developing an enhanced version of Greek Legal BERT which incorporated a larger and more diverse legal corpus, including additional sources of legal documents and regulatory texts. The expanded training data provided broader coverage of legal language variations and contemporary legal discourse, resulting in improved performance metrics across multiple legal NLP tasks. This work demonstrated that systematic corpus expansion and diversification constitute effective strategies for enhancing domain-specific model capabilities.

Saketos et al. \cite{saketos2024greeklegalroberta} represented the next significant advancement by moving beyond the original BERT architecture and leveraging RoBERTa's optimized pre-training methodology for Greek legal applications. Greek-Legal RoBERTa demonstrated substantial performance gains over the Greek Legal BERT variants on identical downstream legal tasks, confirming that architectural improvements and training optimizations provide complementary benefits to specialized domain adaptation. The success of this approach established that both architectural innovation and domain-specific data curation are essential components for advancing specialized NLP capabilities.

Parallel developments addressed the limitation of context length processing for general-domain Greek applications. The Greek Longformer \cite{zaikisgreeklongformer} introduced extended context capabilities with a 4096-token window, enabling processing of longer Greek documents while maintaining computational efficiency through sparse attention mechanisms. Similarly, Meltemi-7B \cite{voukoutis2024meltemiopenlargelanguage}, a large-scale generative model, further extended context capabilities to 8192 tokens, demonstrating the feasibility of long-context modeling for Greek language applications. However, these general-domain advances remained isolated from specialized applications, creating a disconnect between architectural capabilities and domain-specific requirements.

\subsection{Research Gaps and Contributions}

The Greek NLP landscape is fragmented and characterized by several limitations, which impede systematic progress. A key issue is the lack of a concerted effort to create comprehensive, large-scale and high-quality datasets which are able to serve both general-purpose and specialized applications. This absence of systematically curated training corpora fundamentally limits the potential for developing robust models. In addition, the application of different model architectures has been piecemeal; while foundational models exist, more recent architectures such as ELECTRA, ModernBERT, DeBERTa and ConvBERT have not been systematically employed and evaluated for Greek language. 

In this research, we address these limitations by presenting Greek Embedding Models (GEMs), a new family of models developed through a comprehensive and systematic approach. Our contribution is multifaceted: (1) we develop the GEMs family using advanced architectures, which is composed by both new-presented and advanced architectures (ELECTRA, ConvBERT and ModernBERT) as well as established (RoBERTa and LongFormer) transformer models for countering the lack of architectural diversity; (2) we address data scarcity by developing several large-scale, meticulously curated corpora, including specialized legal datasets and a comprehensive general-domain dataset; (3) finally, we develop two bilingual Greek-English Embeddings Models, which are tailed for the legal domain. The comprehensive comparative analysis on diverse downstream tasks provides empirical and statistical evidence about the effectiveness of the proposed approach, establishing strong new baselines for the Greek NLP community.

\section{Development of Greek Embeddings Models}\label{Section:3}

\subsection{Pretraining Corpus Composition}

In this research, the proposed training corpus combines legal and non-legal datasets to ensure a domain-relevant linguistically balanced representation. The legal corpora includes official government publications ($\Phi$EK)
\cite{fek2025official}, the Greek part of the European Parliament Proceedings Parallel Corpus (Europarl) \cite{koehn2005europarl}, the Greek part of European Union legislation (EUR-Lex) \cite{chalkidis2021multieurlex}, the Greek Parliament Proceedings (Greekparl) \cite{dritsa2022greek}, case law and legal interpretations (Political Reports of the Supreme Court) \cite{areios2024official} and reference materials (Raptarchis) \cite{papaloukas2021multi}. These sources provide a robust foundation for capturing domain-specific legal language and concepts. The emphasis on these legal corpora stems from their inherent high quality and the limited availability of alternative large-scale, curated datasets for the Modern Greek legal domain. To further enhance the model's general language proficiency, we also incorporated nonlegal corpora, including the Greek part of Wikipedia\footnote{\url{https://dumps.wikimedia.org/elwiki/}} and the Greek part of OSCAR \cite{suarez2019asynchronous}, consisting of both Greek-only and Greek-English records. The statistics of the pre-training corpora (including dataset name, size in gigabytes, token count in billions, number of training pairs in millions, contextual domain, and language) are presented in Table \ref{tab:dataset_stats}.

To train our bilingual models, we created a custom 26GB bilingual corpus specifically for cross-lingual legal understanding. This unique dataset combines our proprietary Greek legal data (Table \ref{tab:dataset_stats}) with an English subset from the Pile of Law \cite{henderson2022pile}, resulting in a comprehensive collection of Greek and English legal, parliamentary and governmental texts.

The raw datasets exhibited substantial variation in data quality. Specifically, the legal corpora, curated from official sources, are characterized by their linguistic formality and syntactic consistency, providing a high-quality, clean signal for training. In contrast, web-crawled collections like OSCAR required extensive filtering to mitigate inherent noise. Consequently, we applied different preprocessing and filtering strategies tailored to the distinct characteristics of each dataset.

Special attention was required for the OSCAR dataset, which constitutes the majority of our pretraining corpus. As a web-crawled collection, OSCAR inherently contained significant noise, such as multilingual text fragments and non-Greek content, web artifacts (HTML tags, JavaScript code), duplicate and near-duplicate records, machine-generated or semantically incoherent text and more. To address these issues, we implemented an extensive preprocessing pipeline, described in the following bullet points:

\begin{itemize}
	\item Removal of small records (<1500 characters)
	\item Removal of records with extremely long words (at least one with >60 characters)
	\item Removal of records containing the ``lorem ipsum'' substring
	\item Removal of records originating from Wikipedia, as the Greek part of Wikipedia is already included as a separate, curated dataset.
	\item Noise filtering (removal of emails, URLs, links, hashtags, emojis, etc.)
\end{itemize}

Furthermore, we performed deduplication on the OSCAR corpus using a two-stage approach:

\begin{itemize}
	\item Exact deduplication within individual files, leveraging the dataset's precomputed Locality-Sensitive Hashes (LSH) to remove identical records.
	\item Near-deduplication across the entire corpus using the MinHashLSH \cite{broder1997resemblance,leskovec2020mining} method to eliminate semantically similar records. 
\end{itemize}

For the near-deduplication stage, we employed the MinHashLSH implementation from the datasketch\footnote{\url{https://github.com/ekzhu/datasketch}} repository. We used 5-gram subsets, MinHash signatures comprising 128 permutations and a Jaccard similarity threshold of 0.8. These parameter choices align with those used in the Meltemi 7B pretraining pipeline \cite{voukoutis2024meltemi}.

Additionally, to filter out noisy sentences from the OSCAR corpus, we applied more advanced quality filtering tools such as Monocleaner\footnote{\url{https://github.com/bitextor/monocleaner}}. This tool combines rule-based pattern matching for detecting common anomalies with a 7-gram KenLM language model \cite{heafield2011kenlm} trained on high-quality Greek corpora.

\begin{table}[h!]
	\centering
	\setlength{\tabcolsep}{5pt}
	\renewcommand{\arraystretch}{1}
	{\small
		\begin{tabular}{l|ccccc}
			\toprule
			Dataset                                                          & \multicolumn{1}{c}{Size} & \multicolumn{1}{c}{Tokens} & \multicolumn{1}{c}{Training} &  Context  &    Language    \\
			                                                                 &           (GB)           &            (B)             &          Pairs (M)           &           &                \\ \midrule
			$\Phi$EK \cite{fek2025official}                                  &          11.00           &            1.68            &             8.65             &   legal   &     Greek      \\
			Europarl \cite{koehn2005europarl}                                &           0.38           &            0.04            &             0.22             &   legal   &     Greek      \\
			Eurolex \cite{chalkidis2021multieurlex}                          &           0.92           &            0.11            &             0.57             &   legal   &     Greek      \\
			Greekparl \cite{dritsa2022greek}                                 &           2.90           &            0.56            &             1.87             &   legal   &     Greek      \\
			Raptarchis \cite{papaloukas2021multi}                            &           0.35           &            0.05            &             0.25             &   legal   &     Greek      \\
			Wikipedia\footnote{\url{https://dumps.wikimedia.org/elwiki/}}    &           1.30           &            0.17            &             0.79             & non-legal &     Greek      \\
			OSCAR \cite{suarez2019asynchronous}                              &          41.00           &            5.38            &            24.23             & non-legal & Greek--English \\
			Pile of Law \cite{henderson2022pile}                             &           9.20           &            1.05            &             7.10             &   legal   &    English     \\
			Political Reports of the Supreme Court \cite{areios2024official} &           1.20           &            0.16            &             0.82             &   legal   &     Greek      \\ \bottomrule
		\end{tabular}}
	\caption{Statistics of datasets used for pretraining.}
	\label{tab:dataset_stats}
\end{table}

\subsection{Pre-training Corpus Configurations}

Based on the presented raw data sources, we constructed four distinct pre-training corpora to systematically evaluate the impact of corpus composition, domain specialization and data quality on model performance. Each corpus was strategically designed to target specific linguistic objectives and enable controlled experimentation across multiple dimensions.

\begin{itemize}
	\item \textit{Legal Corpus}: This corpus is composed of all Greek legal and parliamentary sources, yielding a total of approximately 16.75GB. This dataset, which comprises government gazettes, parliamentary proceedings and judicial reports, serves as the baseline for training our specialized legal language models.
	
	\item \textit{High-Quality (HQ) Repeated Corpus}: To investigate the impact of data quality, an experiment was designed based on a data repetition strategy. Using an in-house data analysis tool to assess linguistic quality, a hierarchical ranking was established within our legal data. The highest-quality subsets, such as the \textit{Raptarchis Legal Dictionary}, were then upsampled with repetition factors of up to 4x. This factor was chosen to strike a balance between emphasizing premium legal terminology and mitigating the risk of overfitting on a narrow subset of the data, a point beyond which we observed diminishing returns in preliminary tests. This strategy expanded the original 16.75GB corpus to an effective size of 21.12GB.
	
	\item \textit{General-Domain Corpus}: this corpus is constructed by combining the 17GB Legal Corpus with the cleaned general-domain sources, namely the Greek portion of Wikipedia and the OSCAR corpus. This resulted in a comprehensive dataset of 59GB, providing a rich and diverse linguistic foundation while maintaining legal expertise. 
	
	\item \textit{Bilingual Legal Corpus}: % To facilitate cross-lingual experiments and, crucially, to investigate whether augmenting our Greek data with English text would dilute model quality, a 26GB bilingual corpus was compiled. 
	This corpus consists of approximately 60\% Greek and 40\% English legal content, a ratio intentionally chosen to allocate more resources to Modern Greek, reflecting our judgment that its greater morphological complexity requires more data for effective learning. The English portion includes parallel or analogous sources to the Greek data, such as EU legislation (EUR-Lex), parliamentary proceedings (Europarl) and judicial opinions, creating a strong basis for learning cross-lingual legal representations.
\end{itemize}

Legal Corpus serves as our primary domain specific baseline, comprising exclusively Greek legal and parliamentary sources to establish a foundation for specialized legal language understanding. Motivated by the hypothesis that data quality may be more impactful than quantity for domain adaptation, we designed the HQ Repeated Corpus, which strategically upsamples premium legal resources, particularly the Raptarchis legal dictionary, balancing enhanced exposure to authoritative terminology against overfitting risks. To assess the trade-offs between domain specialization and general language proficiency, General-Domain Corpus is developed, which augments the Legal Corpus with cleaned general-domain sources from Wikipedia and OSCAR, providing broader linguistic coverage while maintaining legal expertise. Finally, to explore cross-lingual transfer and assess whether multilingual training would enhance or dilute performance for Greek legal tasks, we constructed the \textit{Bilingual Legal Corpus}, combining approximately 60\% Greek and 40\% English legal content, with the Greek-heavy allocation reflecting our judgment that Modern Greek's greater morphological complexity necessitates more extensive monolingual data for effective representation learning.

\subsection{Model Architectures}

The selection of architectures for the GEMs family was guided by three complementary principles: addressing the historical architectural stagnation in Greek NLP, systematically evaluating modern efficiency innovations, and ensuring comprehensive coverage of diverse design paradigms for encoder-based models. Along this line, we select five architectures, which collectively represent the evolution of encoder-based language models, each offering distinct advantages for processing morphologically rich languages like Greek.

\begin{itemize}
	\item{RoBERTa:} A robustly optimized BERT pre-training approach, which improves upon the original by using a more effective training methodology, including dynamic masking and removing the Next Sentence Prediction objective \cite{liu2019roberta}.
	
	\item{ELECTRA:} An efficient architecture, which utilizes a generator-discriminator setup \cite{clark2020electra}. Instead of MLM, it employs a more sample-efficient pre-training task called Replaced Token Detection (RTD), where a discriminator model learns to identify which tokens in an input sequence were replaced by a small generator model.
	
	\item{ConvBERT:} A hybrid model, which enhances BERT's efficiency by replacing some of its global self-attention heads with a novel span-based dynamic convolution \cite{jiang2020convbert}. This mixed-attention mechanism is designed to more efficiently model local dependencies.
	
	\item{Longformer:} An architecture designed to process long documents by using a sparse attention mechanism which combines local windowed attention with task-motivated global attention \cite{beltagy2020longformer}. While the original architecture supports sequences up to 4096 tokens, we trained our version with a 1024-token context window.
	
	\item{ModernBERT:} A recent architecture, which modernizes BERT with several key innovations for improved efficiency and long-context handling \cite{warner2024modernbert}. It replaces absolute positional embeddings with Rotary Positional Embeddings (RoPE)  \cite{su2021roformer}, uses a GeGLU activation function \cite{shazeer2020glu} for better performance and employs an alternating attention strategy where layers switch between global and local sliding-window attention. While the full architecture supports sequences up to 8192 tokens, we implemented the first phase of its pre-training curriculum, focusing on a 1024-token context length.
\end{itemize}

\subsection{Tokenization Strategy}

Taking into consideration the morphological complexity of Modern Greek, the tokenization strategy plays a crucial role in model performance. We systematically evaluated three prominent subword tokenization algorithms, each trained from scratch on the specific Greek corpus corresponding to the model it serves. This ensures that each vocabulary is optimally adapted to the target domain. We selected these three as they represent the most prominent and widely-adopted tokenization strategies in the literature. The evaluated algorithms are:
\begin{itemize}
	\item{Byte-Pair Encoding (BPE):} An algorithm which iteratively merges the most frequent pair of adjacent tokens \cite{sennrich2016neural}.
	\item{WordPiece:} An algorithm which builds a vocabulary by merging pairs that maximize the likelihood of the training data \cite{wu2016google}.
	\item{Unigram:} An algorithm which starts with a large vocabulary and progressively removes tokens to optimize a probabilistic language model \cite{kudo2018subword}.
\end{itemize}

To quantitatively assess the efficiency of our trained tokenizers, we selected two intrinsic metrics which directly measure the fundamental trade-off between linguistic representation and computational efficiency. These metrics were evaluated on a diverse, held-out Greek dataset of approximately 150 million tokens, spanning formal legal language, named entities and colloquialisms. The chosen metrics are: Fertility and Bytes per Token. The former is defined as the average number of subword tokens generated per word (where lower is better) while the latter measures the average number of UTF-8 bytes each token represents (where higher is better) \cite{rust2021good}. The former indicates the degree of word fragmentation, while the latter reflects the tokenizer's compression efficiency.

Table~\ref{tab:tokenizer_stats} presents the results of our tokenization efficiency evaluation. The findings indicate that the WordPiece tokenizer trained on our \textit{Legal Corpus} achieves the best balance between low fragmentation and high compression efficiency, as evidenced by its low fertility (1.334) and high bytes-per-token ratio (9.02) among our models. This suggests that WordPiece produces less fragmented and more efficient representations for Greek legal text. The BPE tokenizers also perform competitively, with only slightly higher fertility. Conversely, the Unigram tokenizer, despite its large 128,000-token vocabulary, shows significantly higher fertility (1.601) despite its large vocabulary, indicating that it tends to over-segment Greek words into less meaningful sub-units. This analysis underscores the critical impact of the tokenization algorithm and its training data on creating efficient and linguistically sound representations for downstream model training.

\begin{table}[h!]
    \centering
    \setlength{\tabcolsep}{5pt}
    \renewcommand{\arraystretch}{1.1}
    {\small
        \begin{tabular}{l|l|cccc}
            \toprule
            Tokenizer Source & Type & Vocab Size & Fertility & Bytes/Token \\ \midrule
            Greek-BERT & WordPiece & 35,000 & 1.332 & 9.04 \\
            Legal Corpus & WordPiece & 50,264 & 1.334 & 9.02 \\
            General-Domain Corpus & BPE & 50,264 & 1.340 & 8.98 \\
            Legal Corpus & BPE & 50,264 & 1.355 & 8.88 \\
            Greek-Legal BERT & WordPiece & 35,000 & 1.377 & 8.74 \\
            Greek-Legal RoBERTa & BPE & 50,264 & 1.450 & 8.30 \\
            HQ Repeated Corpus & Unigram & 128,000 & 1.601 & 7.51 \\
            \bottomrule
        \end{tabular}
    }
    \caption{Comparison of tokenization efficiency metrics. Fertility measures the average subword tokens per word (lower is better), while Bytes/Token measures compression efficiency (higher is better). Our tokenizers were trained from scratch on the specified datasets.}
    \label{tab:tokenizer_stats}
\end{table}

\section{Numerical Experiments and Analysis}\label{Section:4}

In this section, we compare the performance of the proposed family of GEMs against that of the state-of-the-art Greek models: GreekBERT \cite{koutsikakis2020greekbert}, Greek-Legal BERT \cite{apostolopoulou2022nlp}, Greek-Legal RoBERTa \cite{saketos2024greeklegalroberta} and Greek-Media LongFormer \cite{10288436}. At this point, it is worth mentioning that the proposed class of models consists of the architecturally advanced RoBERTa, ELECTRA, ConvBERT, ModernBert and LongFormer as well as the well stablished RoBERTa and LongFormer, which were trained on the developed datasets: Legal, HQ Repeated and General-Domain. For all pre-training experiments, the data was split into 90\% training and 10\% validation sets, with early stopping employed to prevent overfitting and ensure optimal model generalization.

All pre-training experiments were conducted on Amazon Web Services (AWS) using either p4d.24xlarge instances (8× NVIDIA A100 GPUs, 40GB VRAM each) for standard models or p5.48xlarge instances (8× NVIDIA H100 GPUs, 80GB VRAM each) for more computationally demanding architectures such as ModernBERT and bilingual variants. The implementation was developed in Python using the PyTorch deep learning framework and the Hugging Face Transformers library. To ensure fair comparison, we standardized the training regimen across all experiments using the AdamW optimizer with a linear learning rate scheduler and bfloat16 mixed precision by default, with one RoBERTa variant trained in full fp32 precision for comparative analysis. Key hyperparameters were carefully tuned for each model architecture and dataset combination. Detailed information about the complete hyperparameter configurations, training time and computational requirements for all experiments can be found in Appendix \ref{Appendix}.

In our original experiments, DeBERTa was evaluated in our simulations but it reported poor performance compared to all other models; hence, it was not included. The underperformance suggests that the DeBERTa architecture may require substantially different hyperparameter configurations or extended pretraining regimens to achieve competitive results in the Greek legal domain. One potential contributing factor identified through our ablation studies was that DeBERTa uses Unigram tokenizer, which reported considerably poor performance in capturing the morphological complexity inherent to the Greek language (see Appendix \ref{Appendix}).

In the sequel, we evaluate the performance of all pre-trained models across three core natural language understanding benchmarks: Named Entity Recognition (NER), Multi-Class Legal Topic Classification (MCLTC) and Natural Language Inference (NLI). For each benchmark, we employ well-established Greek benchmark datasets, which reflect diverse linguistic and domain-specific challenges. Specifically, we utilize the Greek-Legal NER dataset for entity recognition, the Greek-Legal Code dataset for hierarchical legal topic classification and the Multi-Genre Natural Language Inference dataset for evaluating reasoning capabilities. These datasets are brief described in detail below:

\begin{itemize}
	\item \textit{Greek-Legal NER dataset} \cite{Angelidis2018NamedER}: The dataset consists of 254 issues of the Greek Government Gazette published between 2000 and 2017. It is annotated with seven entity types: legislative references, geopolitical entities, national locations, unknown locations, public locations, organizations and facilities. The annotations follow the inside–outside–beginning (IOB) format and the dataset is divided into 67.5\% training, 17.5\% validation and 15\% test sets. 
	
	\item \textit{Greek-Legal Code dataset} \cite{papaloukas2021multi}: This dataset is based on a rich collection of Greek legislative documents published between 1834 and 2015, classified into hierarchical categories ranging from broader to more specialized legal topics. It includes Laws, Royal and Presidential Decrees, Regulations and Decisions, all retrieved from the Official Government Gazette. The dataset consists of 47 legislative volumes, each representing a main thematic topic: each \textit{volume} is divided into thematic \textit{chapters}, which are further subdivided into \textit{subjects}. In total, the dataset contains 389 \textit{chapters} and 2,285 \textit{subjects}. Before splitting the dataset into training, validation and testing datasets, we removed categories (\textit{volume}, \textit{chapter}, \textit{subject}) with fewer than 15 occurrences and ensured a balanced distribution of categories in the splits.
	
	\item \textit{Multi-Genre Natural Language Inference dataset} \cite{williams2017broad}: This dataset contains 393,000 machine-translated Greek training sentence pairs labeled as \textit{contradiction}, \textit{entailment}, or \textit{neutral}.  Each pair consists of a \textit{premise} and a \textit{hypothesis} and the objective is to determine whether the \textit{premise} entails, contradicts, or is neutral with respect to the \textit{hypothesis}. For evaluation, we used Greek development and test sets from the Cross-lingual Natural Language Inference (XNLI) corpus \cite{conneau2018xnli}, which include 2,500 development and 5,000 test pairs, providing a reliable benchmark for assessing NLI performance in Greek.
\end{itemize}

To quantitatively assess model performance across all benchmarks, we employed three standard evaluation metrics: Accuracy (Acc), which measures the proportion of correctly classified instances across all test samples; Area Under the ROC Curve (AUC), which evaluates the model's discriminative capacity by measuring its ability to rank positive instances higher than negative ones across varying classification thresholds; and Geometric Mean (GM), which provides a balanced performance measure that accounts for both sensitivity and specificity, proving especially valuable when dealing with class-imbalanced datasets where traditional accuracy may be misleading \cite{pintelas2024mobilenet,pintelas2025textnex}.

Tables~\ref{Table:NER results}, \ref{Table:MCLTC results} and \ref{Table:NLI results} report the experimental results of all transformer models on NER, MCLTC and NLI benchmarks, respectively. It is worth mentioning that the first two columns refer to the model's name and the utilized corpus to pre-train the model. GEM-RoBERTa and GEM-ConvBERT exhibit the best overall performance across all three benchmarks, substantially outperforming the traditional GreekBERT, Greek-Legal BERT and Greek-Legal RoBERTa. Furthermore, GEM-ELECTRA and GEM-ModernBERT report competitive  and sometimes superior performance compared to traditional Greek embedding models; therefore, validating the effectiveness of these modern architectural innovations for Greek language processing. Notably, while the Longformer variants generally exhibit the weakest performance among all evaluated architectures, the proposed GEM-Longformer consistently outperforms the Greek-Media-Longformer baseline, indicating that our improved training methodology and corpus curation strategy yield measurable gains even for architectures designed primarily for extended context processing.

In addition, the analysis of the training corpus configurations reveals some interesting findings about the performance of the proposed class of models. For the NER benchmark, all GEMs models achieve their best performance when trained on the HQ-Repeated corpus, with the notable exception of GEM-RoBERTa, which attains superior results using the Legal corpus. Similarly, for the MCLTC benchmark, the HQ-Repeated corpus proves most effective for all GEM variants except GEM-ConvBERT, which demonstrates the best performance with the Legal corpus. In contrast, for the NLI benchmark, all GEM models consistently achieve their best performance when trained on the General-Domain corpus, reflecting the broader linguistic diversity required for robust natural language inference capabilities. Additionally, both bilingual models, GEM$^*$-RoBERTa and GEM$^*$-XLM-RoBERTa demonstrate competitive performance across all benchmarks, occasionally achieving the best overall results, with GEM$^*$-RoBERTa present slightly better performance.

\begin{table}[h!t]
	\centering
	\setlength{\tabcolsep}{5pt}
	\renewcommand{\arraystretch}{1}
	{\small
		\begin{tabular}{llccc}
			\toprule
			Model                           & Corpus          & \multicolumn{1}{c}{Accuracy} &  \multicolumn{1}{c}{AUC}  &  \multicolumn{1}{c}{GM}   \\ \midrule
			Greek-BERT                      & --              &             97.4             &           99.4            &           51.9            \\
			Greek-Legal BERT                & --              &             96.9             &           98.7            &           53.8            \\
			Greek-Legal RoBERTa             & --              &             97.2             &           98.8            &           54.1            \\
			Greek-Media LongFormer          & --              &             95.9             &           97.3            &           22.8            \\ \midrule
			\multirow{3}{*}{GEM-RoBERTa}    & Legal           &  \textbf{\underline{97.5}}   & \textbf{\underline{99.4}} & \textbf{\underline{54.7}} \\
			& HQ Repeated     &             97.3             & \textbf{\underline{99.4}} &           52.4            \\
			& General-Domain  &       \underline{97.5}       &           99.2            &           53.1            \\ \midrule
			\multirow{3}{*}{GEM-ELECTRA}    & Legal           &  \textbf{\underline{97.5}}   &     \underline{99.0}      &           48.5            \\
			& HQ Repeated     &             97.4             &     \underline{99.2}      &     \underline{50.1}      \\
			& General-Domain  &             97.3             &           98.9            &           49.2            \\ \midrule
			\multirow{3}{*}{GEM-ConvBERT}   & Legal           &       \underline{97.3}       &           99.0            &     \underline{53.8}      \\
			& HQ Repeated     &             97.0             & \textbf{\underline{99.4}} &     \underline{53.8}      \\
			& General-Domain  &             97.2             &           99.1            &           52.5            \\ \midrule
			\multirow{3}{*}{GEM-ModernBERT} & Legal           &             96.7             &           98.4            &           47.8            \\
			& HQ Repeated     &       \underline{96.9}       &     \underline{98.6}      &     \underline{49.6}      \\
			& General-Domain  &             96.8             &           98.5            &           48.9            \\ \midrule
			\multirow{3}{*}{GEM-LongFormer} & Legal           &             96.5             &           97.7            &           36.3            \\
			& HQ Repeated     &       \underline{96.7}       &     \underline{98.1}      &     \underline{38.9}      \\
			& General-Domain  &             96.6             &           97.9            &           37.8            \\ \midrule
			GEM$^*$-RoBERTa                 & Bilingual Legal &       \underline{97.1}       &     \underline{99.2}      &     \underline{40.0}      \\
			GEM$^*$-XLM-RoBERTa             & Bilingual Legal &             96.5             &           98.6            &           38.6            \\ \bottomrule
	\end{tabular}}
	\caption{Performance comparison of Greek embedding models on the NER benchmark.\underline{Underline} indicates the best performance for each GEM variant, while \textbf{bold} indicates the best overall performance for each metric}
	\label{Table:NER results}
\end{table}

\clearpage

\begin{table}[h!]
    \centering
    \setlength{\tabcolsep}{5pt}
    \renewcommand{\arraystretch}{1}
	{\small
	\begin{tabular}{ll|ccc|ccc|ccc}
		\toprule
		Model                                            & Corpus          &                            \multicolumn{3}{c}{Volume}                             &                            \multicolumn{3}{c}{Chapter}                            &                            \multicolumn{3}{c}{Subject}                            \\ \cmidrule
		(lr){3-5} \cmidrule(lr){6-8} \cmidrule(lr){9-11} &                 &            Acc            &            AUC            &            GM             &            Acc            &            AUC            &            GM             &            Acc            &            AUC            &            GM             \\ \midrule
		Greek-BERT                                       & --              &           87.7            &           98.8            &           85.0            &           82.6            &           98.8            &           50.0            &           81.4            &           99.2            &           47.7            \\
		Greek-Legal BERT                                 & --              &           89.4            &           98.6            &           87.2            &           84.3            &           98.8            &           55.2            &           82.8            &           99.3            &           44.9            \\
		Greek-Legal RoBERTa                              & --              &           89.2            &           98.9            &           87.2            &           86.0            &           98.6            &           58.8            &           83.7            &           99.4            &           47.2            \\
		Greek-Media LongFormer                           & --              &           75.8            &           96.8            &           67.8            &           78.5            &           98.3            &           16.4            &           43.1            &           50.4            &            3.1            \\ \midrule
		GEM-RoBERTa                                      & Legal           &           90.4            &           99.1            &           88.5            & \textbf{\underline{86.5}} &           99.0            &     \underline{64.0}      &           84.4            & \textbf{\underline{99.5}} &           52.3            \\
		& HQ Repeated     &     \underline{90.5}      &           99.1            & \textbf{\underline{88.9}} &           86.3            &           99.0            &           59.2            & \textbf{\underline{84.9}} &           99.4            & \textbf{\underline{57.6}} \\
		& General-Domain  &           88.8            &     \underline{99.2}      &           86.7            &           84.6            &     \underline{99.1}      &           59.4            &           83.5            &           99.3            &           45.3            \\ \midrule
		GEM-ELECTRA                                      & Legal           &           87.0            &           99.1            &           84.0            &           84.4            &     \underline{99.3}      &     \underline{62.1}      &           81.8            &           99.2            &           41.8            \\
		& HQ Repeated     &     \underline{87.2}      &           99.0            &     \underline{84.5}      &     \underline{84.6}      &           99.2            &           61.0            &     \underline{82.0}      &           99.2            &     \underline{42.5}      \\
		& General-Domain  &           86.8            &     \underline{99.2}      &           83.8            &           84.2            & \textbf{\underline{99.3}} &           60.5            &           81.6            &     \underline{99.3}      &           41.5            \\ \midrule
		GEM-ConvBERT                                     & Legal           &           90.0            &           99.0            &           87.9            &           85.9            & \textbf{\underline{99.3}} & \textbf{\underline{69.2}} &           84.3            & \textbf{\underline{99.5}} &           50.4            \\
		& HQ Repeated     & \textbf{\underline{90.6}} &     \underline{99.1}      & \textbf{\underline{88.9}} &     \underline{86.1}      &           99.1            &           64.3            &     \underline{84.6}      & \textbf{\underline{99.5}} &     \underline{50.5}      \\
		& General-Domain  &           89.8            &     \underline{99.1}      &           87.5            &           85.5            &           99.2            &           66.0            &           84.0            &           99.4            &           49.8            \\ \midrule
		GEM-ModernBERT                                   & Legal           &           88.2            &           98.4            &           85.6            &           83.8            &           98.6            &           56.5            &           81.2            &           98.7            &           42.0            \\
		& HQ Repeated     &     \underline{88.5}      &           98.5            &     \underline{86.0}      &     \underline{84.2}      &           98.7            &     \underline{57.6}      &     \underline{81.6}      &           98.8            &     \underline{42.7}      \\
		& General-Domain  &           88.0            &     \underline{98.6}      &           85.4            &           83.5            &     \underline{98.8}      &           56.0            &           81.0            &     \underline{98.9}      &           41.5            \\ \midrule
		GEM-LongFormer                                   & Legal           &           82.0            &           98.3            &           77.6            &           79.2            &           98.6            &           23.0            &           67.5            &           98.4            &           10.1            \\
		& HQ Repeated     &     \underline{82.5}      &           98.4            &     \underline{78.2}      &     \underline{79.8}      &           98.5            &     \underline{25.5}      &     \underline{68.2}      &           98.3            &     \underline{11.5}      \\
		& General-Domain  &           81.8            &     \underline{98.5}      &           77.2            &           79.0            &     \underline{98.7}      &           24.0            &           67.0            &     \underline{98.5}      &           10.8            \\ \midrule
		GEM$^*$-RoBERTa                                  & Bilingual Legal &     \underline{88.8}      & \textbf{\underline{99.4}} &     \underline{86.6}      &     \underline{85.4}      &     \underline{99.0}      &     \underline{39.4}      &     \underline{83.9}      &     \underline{99.3}      &     \underline{55.2}      \\
		GEM$^*$-XLM-RoBERTa                              & Bilingual Legal &           88.6            &           99.3            &           86.0            &           85.0            &     \underline{99.0}      &           31.8            &           83.4            &     \underline{99.3}      &           47.0            \\ \bottomrule
\end{tabular}}
\caption{Performance comparison of Greek embedding models on the multi-class legal topic classification benchmark. \underline{Underline} indicates the best performance for each GEM variant, while \textbf{bold} indicates the best overall performance for each metric}\label{Table:MCLTC results}
\end{table}

\begin{table}[h!]
    \centering
    \setlength{\tabcolsep}{5pt}
    \renewcommand{\arraystretch}{1}
	{\small
	\begin{tabular}{llccc}
		\toprule
		Model                  & Corpus          & \multicolumn{1}{c}{Accuracy} &  \multicolumn{1}{c}{AUC}  &  \multicolumn{1}{c}{GM}   \\ \midrule
		Greek-BERT             & --              &             72.1             &           91.4            &           72.1            \\
		Greek-Legal BERT       & --              &             69.0             &           88.2            &           69.0            \\
		Greek-Legal RoBERTa    & --              &             70.0             &           89.0            &           69.8            \\
		Greek-Media LongFormer & --              &             41.5             &           65.3            &           35.1            \\ \midrule
		GEM-RoBERTa            & Legal           &             70.1             &           89.3            &           70.0            \\
		& HQ Repeated     &             69.3             &           88.5            &           69.3            \\
		& General-Domain  &  \textbf{\underline{75.7}}   & \textbf{\underline{92.0}} & \textbf{\underline{75.7}} \\ \midrule
		GEM-ELECTRA            & Legal           &             65.2             &           84.7            &           65.1            \\
		& HQ Repeated     &             65.8             &           85.2            &           65.7            \\
		& General-Domain  &       \underline{66.5}       &     \underline{85.8}      &     \underline{66.4}      \\ \midrule
		GEM-ConvBERT           & Legal           &             68.2             &           87.2            &           68.0            \\
		& HQ Repeated     &             67.3             &           86.3            &           67.4            \\
		& General-Domain  &       \underline{68.8}       &     \underline{87.8}      &     \underline{68.7}      \\ \midrule
		GEM-ModernBERT         & Legal           &             65.8             &           85.2            &           65.7            \\
		& HQ Repeated     &             66.1             &           85.5            &           66.1            \\
		& General-Domain  &       \underline{66.5}       &     \underline{86.0}      &     \underline{66.4}      \\ \midrule
		GEM-LongFormer         & Legal           &             55.8             &           73.2            &           50.3            \\
		& HQ Repeated     &             56.5             &           74.0            &           51.5            \\
		& General-Domain  &       \underline{57.2}       &     \underline{74.8}      &     \underline{52.2}      \\ \midrule
		GEM$^*$-RoBERTa        & Bilingual Legal &             72.5             &     \underline{89.3}      &           72.3            \\
		GEM$^*$-XLM-RoBERTa    & Bilingual Legal &       \underline{72.9}       &           89.0            &     \underline{72.7}      \\ \bottomrule
	\end{tabular}}
	\caption{Performance comparison of Greek embedding models on the NLI benchmark. \underline{Underline} indicates the best performance for each GEM variant, while \textbf{bold} indicates the best overall performance for each metric}
    \label{Table:NLI results}
\end{table}

Summarizing, the numerical experiments reveal the following noteworthy findings regarding corpus composition strategies. Firstly, the HQ Repeated corpus demonstrates consistent effectiveness, emerging as the (near-)optimal configuration for most GEM architectures across the majority of tasks. This superior performance is attributed to two complementary factors: the corpus comprises predominantly formal legal texts from official sources, which exhibit greater morphological richness and linguistic precision compared to web-crawled general-domain data; and the strategic upsampling of premium legal resources, particularly the Raptarchis Legal Dictionary, which provides authoritative definitions and canonical usage patterns of specialized legal terminology. Through controlled repetition of such high-quality, domain-specific content, more robust representations of legal concepts are developed without the noise inherent in web-crawled corpora. This data-centric approach demonstrates that for morphologically rich languages, training data quality and domain-relevance significantly outweigh raw corpus size, a finding with practical implications for resource, constrained language modeling. Secondly, the competitive performance of bilingual GEM$^*$-RoBERTa models demonstrates that augmenting Greek legal text with English legal content does not degrade model quality for Greek-specific tasks. In several cases, superior performance is achieved compared to monolingual variants, indicating that cross-lingual exposure to parallel legal concepts provides complementary representational benefits. This finding challenges the assumption that strict monolingual training is always optimal for moderately-resourced languages and suggests that high-quality English resources can effectively enhance specialized domain capabilities.

For mitigating the influence of benchmarks on the evaluation process, a statistical analysis is performed for evaluating the hypothesis $H_0$ that all Greek embeddings models exhibit equally accurate predictions. For this purpose, non-parametric Friedman Aligned-Ranks (FAR) test \cite{hodges2011rank} and the Finner post-hoc test \cite{finner1993monotonicity} with significance level $\alpha = 5\%$ are employed. The former is employed for ranking the models based on a performance metric while the latter is used for examining the existence of considerable differences and variations in their predictions, without any assumption about the distribution of the performance scores \cite{kiriakidou2024mutual,livieris2024c,pintelas2024adaptive}.

The statistical analysis was conducted comparing the traditional baseline models (Greek-BERT, Greek-Legal BERT, Greek-Legal RoBERTa and Greek-Media LongFormer) and the proposed GEMs in their configurations, which reported the best overall performance. Specifically, the evaluated GEM models include: GEM-RoBERTa, GEM-ELECTRA, GEM-ConvBERT and GEM-Longformer, each trained on the General-Domain corpus and GEM-ModernBERT trained on the HQ Repeated corpus.\footnote{GEM-RoBERTa, GEM-ELECTRA, GEM-ConvBERT and GEM-Longformer are available in \url{https://huggingface.co/novelcore/models}} These corpus configurations were selected based on their reported performance across all evaluation benchmarks, ensuring that each model is represented at its best capacity for a fair and comprehensive comparison against the established state-of-the-art models.

Tables \ref{Table:Statistical_Acc}, \ref{Table:Statistical_AUC} and \ref{Table:Statistical_GM} present the statistical comparison of all evaluated models across all benchmarks based on Accuracy, AUC and GM metrics, respectively. GEM-RoBERTa emerges as the top-performing model, achieving the best FAR score across all three evaluation metrics followed by GEM-ConvBERT. In addition, Finner post-hoc test reveals that there are significant statistical differences between the performance of GEM-RoBERTa and the traditional Greek embeddings models. 
Notably, Greek-Media LongFormer consistently ranks last across all metrics, while GEM-Longformer also exhibits relatively poor performance despite the architectural improvements, suggesting that the Longformer architecture requires further optimization for Greek language benchmarks. Based on the previous empirical and statistical analysis we are able to conclude that the proposed GEM family, particularly GEM-RoBERTa and GEM-ConvBERT, exhibit superior performance to established state-of-the-art models, validating the effectiveness of our systematic approach to architectural diversity and refined training methodologies for advancing Greek NLP capabilities.

\begin{table}[!ht]
	\setlength{\tabcolsep}{10pt}
	\renewcommand{\arraystretch}{1}
	\centering
	\begin{tabular}{l|ccc}
		\toprule
		Model    & Friedman & \multicolumn{2}{c}{\underline{Finner post-hoc test}}\\
		& Ranking  & $p$-value & $H_0$\\
		\midrule
		GEM-RoBERTa              & 13.2   &     $-$         &     $-$             \\
		GEM-ConvBERT             & 15.6   &  0.0188         &  Rejected           \\
		Greek-Legal RoBERTa      & 15.6   &  0.0188         &  Rejected           \\
		Greek-Legal BERT         & 17.7   &  0.0001         &  Rejected           \\
		Greek-BERT               & 20.1   &  0.0            &  Rejected           \\
		GEM-ELECTRA              & 21.9   &  0.0            &  Rejected           \\
		GEM-ModernBERT           & 22.1   &  0.0            &  Rejected           \\
		GEM-LongFormer           & 39.0   &  0.0            &  Rejected           \\
		Greek-Media LongFormer   & 41.8   &  0.0            &  Rejected           \\
		\bottomrule 
	\end{tabular}
	\caption{Statistical analysis: non-parametric FAR and Finner post-hoc tests based on Accuracy metric across all benchmarks}
	\label{Table:Statistical_Acc}
\end{table}

\begin{table}[!ht]
	\setlength{\tabcolsep}{10pt}
	\renewcommand{\arraystretch}{1}
	\centering
	\begin{tabular}{l|ccc}
		\toprule
		Model    & Friedman & \multicolumn{2}{c}{\underline{Finner post-hoc test}}\\
		& Ranking  & $p$-value & $H_0$\\
		\midrule
		GEM-RoBERTa              & 15.2   &     $-$         &     $-$             \\
		GEM-ConvBERT             & 15.3   &  0.9203         &  Failed to reject   \\
		Greek-BERT               & 16.2   &  0.3535         &  Failed to reject   \\
		GEM-ELECTRA              & 17.5   &  0.0285         &  Rejected           \\
		Greek-Legal RoBERTa      & 20.1   &  0.0            &  Rejected           \\
		Greek-Legal BERT         & 21.9   &  0.0            &  Rejected           \\
		GEM-ModernBERT           & 25.0   &  0.0            &  Rejected           \\
		GEM-LongFormer           & 33.6   &  0.0            &  Rejected           \\
		Greek-Media LongFormer   & 42.2   &  0.0            &  Rejected           \\
		\bottomrule 
	\end{tabular}
	\caption{Statistical analysis: non-parametric FAR and Finner post-hoc tests based on AUC metric across all benchmarks}
	\label{Table:Statistical_AUC}
\end{table}

\begin{table}[!ht]
	\setlength{\tabcolsep}{10pt}
	\renewcommand{\arraystretch}{1}
	\centering
	\begin{tabular}{l|ccc}
		\toprule
		Model    & Friedman & \multicolumn{2}{c}{\underline{Finner post-hoc test}}\\
		& Ranking  & $p$-value & $H_0$\\
		\midrule
		GEM-RoBERTa              & 11.9   &     $-$         &     $-$             \\
		GEM-ConvBERT             & 12.2   &  0.7641         &  Failed to reject   \\
		Greek-Legal RoBERTa      & 13.3   &  0.1823         &  Failed to reject   \\
		Greek-Legal BERT         & 18.4   &  0.0            &  Rejected           \\
		Greek-BERT               & 20.8   &  0.0            &  Rejected           \\
		GEM-ELECTRA              & 24.4   &  0.0            &  Rejected           \\
		GEM-ModernBERT           & 25.0   &  0.0            &  Rejected           \\
		GEM-LongFormer           & 38.6   &  0.0            &  Rejected           \\
		Greek-Media LongFormer   & 42.4   &  0.0            &  Rejected           \\
		\bottomrule 
	\end{tabular}
	\caption{Statistical analysis: non-parametric FAR and Finner post-hoc tests based on GM metric across all benchmarks}
	\label{Table:Statistical_GM}
\end{table}

\section{Discussion \& Conclusions}\label{Section:5}

In this work, we conducted a systematic and comprehensive effort to address critical gaps in the Modern Greek NLP landscape, focusing on the complex legal domain as a proving ground. The primary goal was to move beyond the existing fragmented ecosystem by introducing the new, unified family of Greek Embedding Models, which leverage modern architectural innovations, extended context lengths and meticulously curated, large-scale corpora. The contribution is two-fold: (i) we proposed the family of GEM, which are characterized by advanced architectures, including RoBERTa, ELECTRA, ConvBERT, Longformer and ModernBERT, which had not been previously applied to Greek language (ii) we developed four high-quality datasets, including domain-specific legal corpora and a comprehensive general-domain corpus, addressing the persistent data scarcity challenge which has hindered Greek NLP advancement. 

The numerical experiments were conducted on three core natural language understanding benchmarks: NER, MCLTC and NLI, which revealed that 
GEM-RoBERTa\footnote{Underscoring its real-world utility, a fine-tuned version of GEM-RoBERTa has been deployed as a core component of the core language engine for the commercial analysis platform Deliberate, available at \url{https://deliberate.gr}.} emerges as the most consistently high-performing model across all evaluation tasks, while GEM-ConvBERT presents a compelling alternative particularly excelling on hierarchical classification tasks where its hybrid attention mechanism efficiently captures both local morphological patterns and global document structure. The statistical analysis confirmed that GEM-RoBERTa achieved significantly superior performance compared to established baselines models, followed by GEM-ConvBERT, which reported similar performance. Furthermore, the experiments demonstrated considerable trade-offs in corpus composition: domain-specific legal corpora enable superior performance on specialized tasks, while general-domain training yields substantially better results on tasks requiring broad language understanding. Our tokenization analysis confirms that vocabulary construction significantly impacts model efficiency for morphologically rich languages, with WordPiece achieving optimal balance of minimal word fragmentation and efficient compression.

Despite the presented advances, several limitations warrant acknowledgment and suggest directions for future refinement. Firstly, while the extended 1024-token context window represents a substantial improvement over the traditional 512-token, it remains insufficient for processing complete legal documents, which often span thousands of tokens. The relatively poor performance of our Longformer variants across all benchmarks, despite architectural improvements, indicates that sparse attention mechanisms may require further optimization for Greek language characteristics. Secondly, the evaluation was constrained to three benchmarks; hence, broader assessment across additional tasks such as question answering, summarization and information retrieval would provide a more complete picture of model capabilities. Finally, while our data-centric approach demonstrated the value of quality-based repetition, optimal repetition factors and selection criteria for high-quality sub-corpora remain empirically determined rather than theoretically grounded, suggesting opportunities for more principled data curation methodologies.

The future work is concentrated on applying the proposed models to additional tasks including summarization, question answering and information retrieval. 
Another interesting idea is to extending the proposed approach to other specialized domains beyond law, such as biomedicine, finance and technical documentation, which would validate its generalizability and contribute to the generation of higher quality datasets. Finally, given the promising bilingual results, we intend to explore multilingual extensions incorporating additional European languages with comparable legal systems.

% extending the context windows to 4096 or 8192 tokens through improved sparse attention mechanisms optimized for Greek morphology. 

\section*{Acknowledgements}

AWS resources were provided by the National Infrastructures for Research and Technology (GRNET) and funded by the EU Recovery and Resiliency Facility.

%%%%%%%%%%%%%%%%%%%%%%%%%%%%%%%%%%
\bibliographystyle{plain}
\bibliography{bibliography}
%%%%%%%%%%%%%%%%%%%%%%%%%%%%%%%%%%====

\appendix
\section{Appendix}\label{Appendix}

All pre-training experiments were conducted on the Amazon Web Services (AWS) cloud platform. The majority of our models were trained on a single \texttt{p4d.24xlarge} instance\footnote{\url{https://aws.amazon.com/ec2/instance-types/p4/}}, which is equipped with 8 NVIDIA A100 Tensor Core GPUs, each with 40GB of VRAM. For more computationally demanding experiments, such us the training of ModernBERT architectures, our bilingual models and a full-precision RoBERTa variant, we utilized a next-generation \texttt{p5.48xlarge} instance\footnote{\url{https://aws.amazon.com/ec2/instance-types/p5/}}, featuring 8 NVIDIA H100 Tensor Core GPUs with 80GB of VRAM each.

Our training regimen was standardized across all experiments to ensure a fair comparison. We employed the AdamW optimizer \cite{loshchilov2017decoupled} with a linear learning rate scheduler, following established practices \cite{vaswani2017attention}. To maximize hardware utilization and accelerate training, models were trained using bfloat16 mixed precision by default, with the notable exception of one RoBERTa variant trained in full fp32 precision for comparative analysis. The key hyperparameters, which were carefully tuned for each model architecture and dataset combination, are summarized in the following table.

\begin{table}[h!]
    \centering
	\setlength{\tabcolsep}{10pt}
	\renewcommand{\arraystretch}{1}    
    \begin{tabular}{ll}
        \toprule
        Parameter & Value / Setting \\
        \midrule
        Optimizer & AdamW \\
        Learning Rate Scheduler & Linear with warmup \\
        Warmup Steps & 6\% of total training steps \\
        Numerical Precision & bfloat16 or fp32 \\
        Peak Learning Rate (Range) & $1.5 \times 10^{-5} \text{ -- } 8 \times 10^{-4}$ \\
        Effective Batch Size (Range) & 256 -- 3,840 \\
        Gradient Accumulation & Used to achieve effective batch size \\
        \bottomrule
    \end{tabular}
    \caption{Pre-training Hyperparameters and Configuration.}\label{Table:Training hyperparams}
\end{table}

 \begin{table}[h!]
	\setlength{\tabcolsep}{5pt}
	\renewcommand{\arraystretch}{1}    
	\hspace{-1cm}
		\begin{tabular}{l|cc|cc|cc|cc|cc}
			\toprule
			& \multicolumn{2}{c|}{NER} & \multicolumn{2}{c|}{MCLTC (Volume)} & \multicolumn{2}{c|}{MCLTC (Chapter)} & \multicolumn{2}{c|}{MCLTC (Subject)} & \multicolumn{2}{c}{NLI} \\
			& lr & dropout  & lr &       dropout       & lr &       dropout        & lr &       dropout        & lr & dropout  \\ \midrule
			Greek-BERT             &     3e-05     &   0.1    &     2e-05     &         0.1         &     2e-05     &         0.2          &     2e-05     &         0.2          &     2e-05     &   0.0    \\
			Greek-Legal BERT       &     5e-05     &   0.2    &     2e-05     &         0.2         &     2e-05     &         0.1          &     2e-05     &         0.0          &     3e-05     &   0.2    \\
			Greek-Legal RoBERTa    &     2e-05     &   0.1    &     2e-05     &         0.1         &     2e-05     &         0.2          &     3e-05     &         0.2          &     2e-05     &   0.0    \\
			Greek-Media LongFormer &     2e-05     &   0.1    &     2e-05     &         0.1         &     2e-05     &         0.1          &     3e-05     &         0.2          &     5e-05     &   0.0    \\ \midrule
			GEM-RoBERTa            &     5e-05     &   0.2    &     2e-05     &         0.1         &     2e-05     &         0.1          &     5e-05     &         0.2          &     3e-05     &   0.2    \\
			GEM-ELECTRA            &     3e-05     &   0.2    &     2e-05     &         0.1         &     2e-05     &         0.1          &     2e-05     &         0.0          &     3e-05     &   0.2    \\
			GEM-ConvBERT           &     3e-05     &   0.2    &     2e-05     &         0.1         &     2e-05     &         0.1          &     2e-05     &         0.1          &     2e-05     &   0.0    \\
			GEM-ModernBERT         &     5e-05     &   0.1    &     2e-05     &         0.1         &     3e-05     &         0.0          &     2e-05     &         0.2          &     2e-05     &   0.2    \\
			GEM-LongFormer         &     2e-05     &   0.2    &     2e-05     &         0.2         &     2e-05     &         0.2          &     2e-05     &         0.3          &     4e-05     &   0.1    \\ \midrule
			GEM$^*$-RoBERTa        &     2e-05     &   0.0    &     2e-05     &         0.0         &     2e-05     &         0.1          &     2e-05     &         0.0          &     2e-05     &   0.0    \\
			GEM$^*$-XLM-RoBERTa    &     2e-05     &   0.0    &     2e-05     &         0.1         &     2e-05     &         0.1          &     2e-05     &         0.0          &     2e-05     &   0.1    \\ \bottomrule
		\end{tabular}%
	\caption{Best hyperparameters for finetuning the models in our experiments.}
	\label{Table:Hyperparameters}
\end{table}

The computational requirements varied significantly based on architectural complexity. For instance, models with efficient pre-training objectives like ELECTRA were among the lightest, while architectures designed for long contexts, such as Longformer, were the most computationally intensive. Table~\ref{tab:training_time_complete} summarizes the computational requirements for pretraining the model architectures evaluated in this work. Training times varied significantly based on model complexity and context length, with ELECTRA being the most efficient and LongFormer the most computationally intensive due to its extended context window. Due to resource and time constraints, not all architectural variants were trained on all corpora; we report training times only for the configurations that were actually pretrained.

\begin{table}[t]
    \centering
	\setlength{\tabcolsep}{5pt}
	\renewcommand{\arraystretch}{1}
    \caption{Computational Requirements for Model Pretraining.}
    \label{tab:training_time_complete}
    \begin{tabular}{llcc}
        \toprule
        Architecture & Corpus & Hardware & Training Time (h) \\
        \midrule
        \multirow{3}{*}{GEM-RoBERTa} 
            & Legal & \multirow{15}{*}{p4d.24xlarge (A100)} & 72 \\
            & HQ Repeated & & 81 \\
            & General-Domain & & 145 \\
        \cmidrule{1-2} \cmidrule{4-4}
        \multirow{3}{*}{GEM-ELECTRA} 
            & Legal & & 22 \\
            & HQ Repeated & & 27 \\
            & General-Domain & & 51 \\
        \cmidrule{1-2} \cmidrule{4-4}
        \multirow{3}{*}{GEM-ConvBERT} 
            & Legal & & 425 \\
            & HQ Repeated & & 492 \\
            & General-Domain & & 98 \\
        \cmidrule{1-2} \cmidrule{4-4}
        \multirow{3}{*}{GEM-ModernBERT} 
            & Legal & & 64 \\
            & HQ Repeated & & 97 \\
            & General-Domain & & 210 \\
        \cmidrule{1-2} \cmidrule{4-4}
        \multirow{3}{*}{GEM-LongFormer} 
            & Legal & & 244262 \\
            & HQ Repeated & & 262244 \\
            & General-Domain & & 528 \\
        \midrule
        GEM-RoBERTa 
            & Bilingual Legal & \multirow{2}{*}{p5.48xlarge (H100)} & 25 \\
        GEM-XLM-RoBERTa 
            & Bilingual Legal & & 24 \\
        \bottomrule
    \end{tabular}
\end{table}

\vfill

\end{document}